\documentclass[]{jingdong}

\usepackage{amsmath,amsfonts,bm}

\def\eqref#1{equation~\ref{#1}}
\def\Eqref#1{Equation~\ref{#1}}
\def\1{\bm{1}}

\DeclareMathAlphabet{\mathsfit}{\encodingdefault}{\sfdefault}{m}{sl}
\SetMathAlphabet{\mathsfit}{bold}{\encodingdefault}{\sfdefault}{bx}{n}

\newcommand{\softmax}{\mathrm{softmax}}

\DeclareMathOperator*{\argmin}{arg\,min}

\usepackage{nicefrac}
\usepackage{enumitem}
\usepackage{mathtools}
\usepackage{amssymb}
\usepackage{wrapfig}
\usepackage{longtable}
\usepackage{float}
\usepackage{placeins}
\DeclareTextFontCommand{\textbf}{\bfseries}

\definecolor{qcodblue}{HTML}{247BA0}
\definecolor{qcodteal}{HTML}{2A9D8F}
\definecolor{qcodnavy}{HTML}{17324D}
\newcommand{\method}{\textsc{QCOC}}
\newcommand{\fullop}{\mathcal{A}_{\mathcal D}}
\newcommand{\compop}{\widehat{\mathcal A}_{\mathcal D}}

\newcommand{\Real}{\mathbb R}

\title{Compile the Table: Query-Calibrated Operator Compression for Tabular In-Context Learning}

\author[1,*]{Xu Zhao}
\author[1,*]{Jiaming Zhao}
\author[1]{Bin Zhao}
\author[1]{Yong Yang}

\affiliation[1]{JD.COM}
\contribution[*]{Equal contribution.}

\checkdata[Email]{\texttt{\{zhaoxu.135, zhaojiaming.214\}@jd.com}}

\hypersetup{
  pdftitle={Compile the Table: Query-Calibrated Operator Compression for Tabular In-Context Learning},
  pdfauthor={Xu Zhao, Jiaming Zhao, Bin Zhao, Yong Yang}
}

\abstract{Tabular in-context learning (ICL) has emerged as a training-free and accurate paradigm for tabular prediction, but current approaches to compressing its in-context examples face an accuracy--throughput tradeoff: fixed subsets can sacrifice accuracy, while query-specific retrieval limits cache reuse and batching across queries, reducing throughput. We propose \method{} (\emph{Query-Calibrated Operator Compression}), which exploits the exchangeability and repeated use of in-context examples by compiling their full KV cache once into compact memory shared across subsequent queries. Instead of retaining raw examples, \method{} clusters their states into joint-KV prototypes, preserves per-cluster multiplicities and the original example count, and calibrates prototype values against attention query vectors produced by the in-context examples through an anchored closed-form solution. Prototype compression drives the speedup, while value fitting helps preserve accuracy. On 64 held-out OpenML-CC18 datasets, \method{} achieves the highest mean accuracy among the compared compression and retrieval methods at both retained counts. Across 12 configurations on seven long tables, it ranks first among compressed methods in ten and averages 0.23 percentage points below full context. Compressing 8,192 in-context examples to 512 memory slots yields a 10.5$\times$ cache compression ratio; excluding one-time compilation, in a single-core CPU online-serving comparison over 1,000 queries, \method{} is up to 508$\times$ faster than dynamic retrieval baselines and 1.98$\times$ faster than full-context inference. These results show that \method{} enables compact-memory reuse and efficient inference across queries while retaining accuracy close to full context.}

\begin{document}

\maketitle

% Begin sections/introduction
\section{Introduction}
\label{sec:introduction}

Tabular in-context models predict new rows from a labeled in-context example set~\citep{tabpfn,tabpfnv2,tabicl}. Larger contexts expose more in-context information, but increase cache memory and repeated-query inference cost~\citep{scalingtabpfn,chunkedtabpfn}. When one in-context example set serves many queries, \emph{what should compression preserve}?

Existing context-compression methods mainly select in-context examples in two ways. \emph{Global row selection} constructs one subset for all queries, making the selected context reusable but often discarding rows needed for accurate prediction~\citep{lucos,feldman2020coresets,wang2018datasetdistillation}. \emph{Dynamic row selection} retrieves a query-specific subset and often improves local relevance, but limits context reuse and batching across queries, thereby reducing serving throughput~\citep{mixturepfn,crumb,arash}. KATE~\citep{kate}, for example, performs nearest-neighbor retrieval in a pretrained representation before each prediction. This creates a central accuracy--throughput tradeoff: global selection has low serving cost but may lose relevant rows, whereas dynamic selection can preserve local relevance at higher serving cost.

To resolve this tradeoff, we propose \emph{Query-Calibrated Operator Compression} (\method{}), which shifts the object of compression from in-context examples to the query--output mapping defined by the in-context example set. For a fixed in-context example set $\mathcal D$, each attention head defines a mapping from a query state $q$ to an attention output $\mathcal A_{\mathcal D}(q)$; we refer to this mapping as the \emph{in-context-example-conditioned attention operator}. Once the in-context example set is fixed, the full KV cache determines this mapping: subsequent queries share the same cache while producing query-dependent outputs. We study tabular compression through \emph{reusable operator compilation}: approximate the full-cache operator once with a compact memory and apply the compiled operator to subsequent queries. Section~\ref{sec:method} gives the formal definition.

Concretely, \method{} compiles the operator induced by an in-context example set into a compact, reusable KV memory. Because the labeled in-context examples are exchangeable, their individual identities and input order need not be retained: at every layer and head, joint-KV clustering groups their states into prototypes, while exact cluster counts restore the total attention contribution of the examples represented by each prototype. Any length-dependent query transformation is evaluated using the original number of in-context examples $N$ rather than the number of compressed slots $M$. Attention query vectors produced by the in-context examples are passed through the full cache to obtain target outputs. With the compressed keys and cluster counts fixed, one regularized least-squares solve fits the compressed value vectors to reproduce these outputs. The compiled KV memory is reused by every future query while still producing a query-specific attention output.

\begin{figure*}[t]
    \centering
    \includegraphics[width=\textwidth]{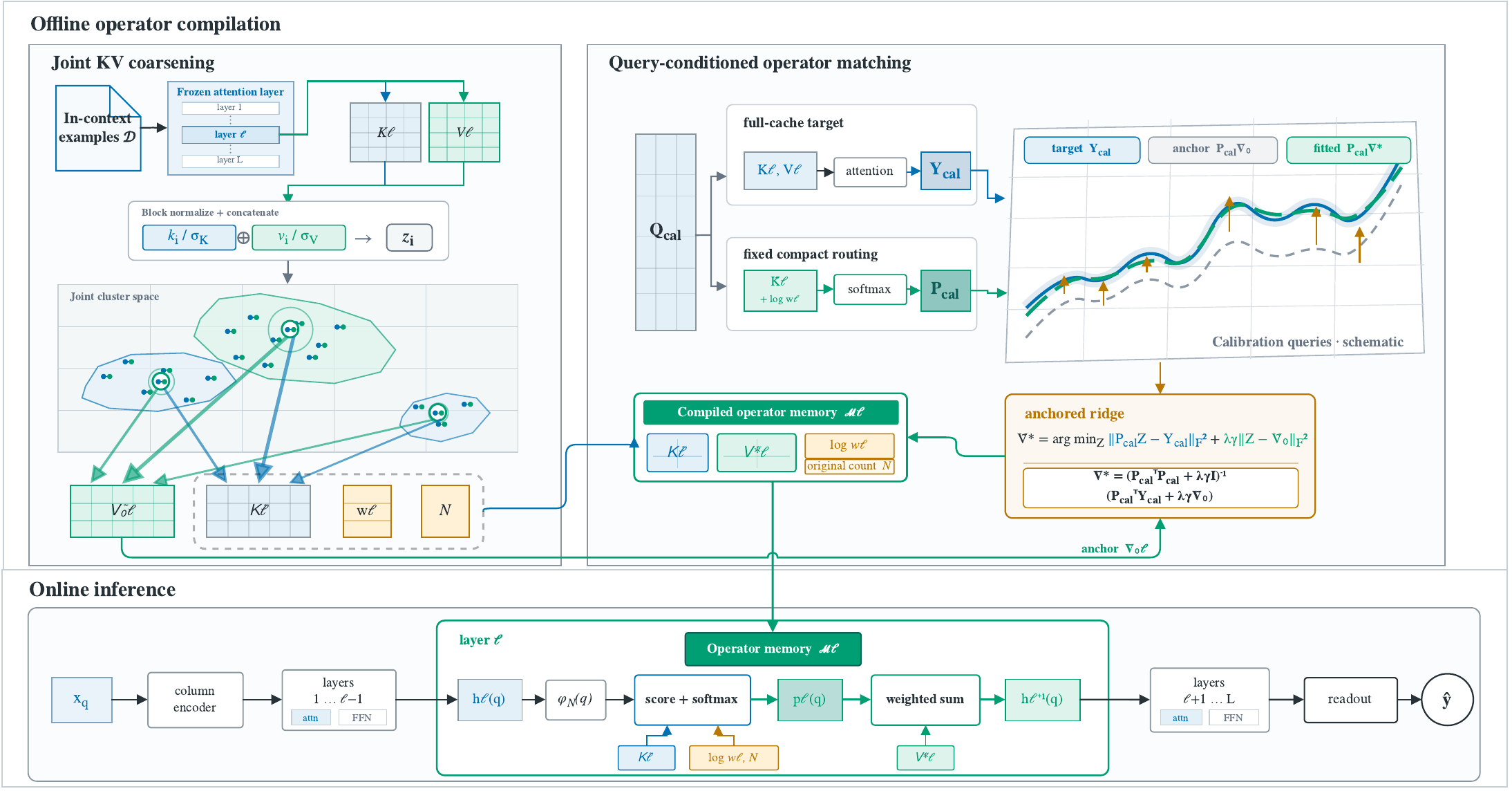}
    \caption{\textbf{Reusable operator compilation with QCOC.} The full in-context cache induces a query-dependent attention operator. Attention query vectors produced by the in-context examples are passed through the full cache to obtain target outputs, and QCOC compresses the cache into $(\widetilde K_\ell,\widetilde V_\ell^*,\log w_\ell,N)$ memory reused by future queries. Joint-KV clustering, cluster-count weighting, and anchored ridge fitting implement this process.}
    \label{fig:overview}
\end{figure*}

Our contributions are:
\begin{itemize}[leftmargin=*,itemsep=2pt,topsep=3pt]
    \item We propose Query-Calibrated Operator Compression (\method{}), a post-hoc KV-compression method for tabular ICL. It uses the exchangeability and repeated reuse of labeled in-context examples to aggregate their states into joint-KV prototypes with exact cluster counts, then fits the prototype values using attention query vectors produced by the examples, producing compressed KV memory reusable across queries without extra data or model updates.
    \item We derive a prototype approximation bound that relates attention-output error to key and value approximation errors, and show that the closed-form value fit cannot increase the residual on its calibration queries. Additional analysis characterizes how approximation error may propagate or change across query distributions.
    \item Through the primary benchmark, long-context evaluation, transfer-model experiments, and multi-query serving experiments, we show that \method{} retains predictive performance close to full context after KV-cache compression and reduces online inference cost when the cache is reused across queries. Component analysis further clarifies the roles of cluster counts, the original example count $N$, and value fitting, while failure-case analysis identifies the method's limitations.
\end{itemize}

% End sections/introduction

% Begin sections/related_work
\section{Related Work}
\label{sec:related}

\paragraph{Tabular ICL and long-context scaling.}
Prior-data fitted networks learn to perform supervised prediction in context~\citep{muller2022pfn,tabpfn}. TabPFN v2 strengthens small-data prediction, while TabICL scales model capacity through a column-then-row architecture~\citep{tabpfnv2,tabicl}. Several works address the remaining in-context example count bottleneck at the input or execution level. Scaling TabPFN studies row sketching and feature selection, BAPS constructs class-balanced prototypes, and Chunked TabPFN preserves exact full-context computation through tiled-block attention~\citep{scalingtabpfn,baps,chunkedtabpfn}. TuneTables and in-context data distillation optimize synthetic contexts or model parameters for a target table~\citep{feuer2024tunetables,xu2024context}. TACO instead trains a dedicated compressor and predictor end to end, mapping the training table into a reusable latent context~\citep{taco}. \method{} differs by post-hoc compiling the internal in-context-example KV states of an existing pretrained tabular ICL model, without updating model parameters or replacing its predictor.

\paragraph{Global and dynamic row selection.}
Global methods select one context that is reused across queries. They include random or clustering-based subsets, LUCoS K-Medoids in a learned tabular representation, and more general coresets or distilled datasets~\citep{lucos,feldman2020coresets,wang2018datasetdistillation}. Their one-time selection supports efficient serving, but a single subset cannot preserve all query-relevant regions. Dynamic methods instead adapt the selected rows to each query: KATE performs nearest-neighbor retrieval, MICP routes queries among contexts, CRUMB constructs distribution-matched query batches, and ARASH adapts retrieval and context size~\citep{kate,mixturepfn,crumb,arash}. Dynamic selection improves local relevance but limits context reuse and batching across queries. \method{} mitigates this tradeoff by compiling one reusable layer-wise KV memory whose attention output remains query-dependent.

\paragraph{Reusable context representations.}
Language-model work has learned to replace long prompts with reusable hidden representations. Gist tokens and ICAE train compact memory tokens, Activation Beacon compresses layer activations directly, and recent evidence shows that even pooling is a strong baseline when evaluated carefully~\citep{gisttokens,icae,activationbeacon,nomeanfeat}. Cartridges and latent context compilation further study reusable memories for later queries~\citep{cartridges,latentcontext}.

\paragraph{KV-cache compression.}
Efficient-attention methods replace quadratic attention with low-rank features or landmarks~\citep{nystromformer,performer}. KV-cache compression has been introduced for autoregressive language models through token eviction~\citep{h2o,scissorhands,snapkv}, state merging~\citep{tome,kvmerger}, and layerwise budget allocation~\citep{pyramidkv}. Recent language-model studies have also explored optimizing or reconstructing compressed KV states beyond token eviction and state merging~\citep{attentionmatching,kvsculpt,grkv,gracekv}.

% End sections/related_work

% Begin sections/formulation
\section{Query-Calibrated Operator Compression}
\label{sec:formulation}
\label{sec:method}

\subsection{Problem Setup and Objective}

Let $\mathcal D=\{(x_i,y_i)\}_{i=1}^{N}$ denote an in-context example set and $f_\theta$ a tabular ICL model whose parameters remain fixed during compression. Encoding $\mathcal D$ produces layer- and head-specific keys and values $K_{\ell h},V_{\ell h}\in\Real^{N\times d}$. Fixing one layer--head pair, write its states as $K,V$.

Given a full KV cache constructed from a labeled in-context example set, our goal is to compress it once into layer- and head-specific KV memory that can be used directly by the predictor and shared across subsequent queries.

For a query state $q\in\Real^d$, define the attention output of the in-context example set as
\begin{equation}
    \fullop(q)=p_{\mathcal D}(q)V,
    \qquad
    p_{\mathcal D}(q)=\softmax\!\left(\phi_N(q)K^\top\right),
    \label{eq:full-operator}
\end{equation}
where $\phi_N(q)$ denotes the query transformation, with the subscript indicating that it may depend on the number of in-context examples. For standard scaled dot-product attention, $\phi_N(q)=q/\sqrt d$ and is independent of $N$. Layer and head indices are suppressed. For an arbitrary batch $Q=[q_1^\top;\ldots;q_B^\top]\in\Real^{B\times d}$, its row-wise extension is
\begin{equation}
    \Phi_N(Q)=[\phi_N(q_1)^\top;\ldots;\phi_N(q_B)^\top]\in\Real^{B\times d}.
    \label{eq:batch-query-map}
\end{equation}
Correspondingly, define $\fullop(Q)=\softmax_{\mathrm{row}}(\Phi_N(Q)K^\top)V$. The cache remains fixed across queries while its output remains query-dependent. Operator compression therefore approximates the function $q\mapsto\fullop(q)$. We represent it with $M\ll N$ keys $\widetilde K\in\Real^{M\times d}$, values $\widetilde V\in\Real^{M\times d}$, and count-based mass weights $w\in\mathbb N^M$:
\begin{align}
    P(Q;\widetilde K,w)
    &=\softmax_{\mathrm{row}}\!\left(\Phi_N(Q)\widetilde K^\top+\mathbf 1_B(\log w)^\top\right),
    \label{eq:compressed-routing}\\
    \compop(Q)&=P(Q;\widetilde K,w)\widetilde V.
    \label{eq:compressed-operator}
\end{align}
The weight $w_j=|D_j|\geq1$ is the number of in-context examples assigned to prototype $j$ in a nonempty partition $D_1,\ldots,D_M$ of the in-context-example states, while $\Phi_N(Q)$ keeps the backbone's query transformation unchanged. For a length-aware backbone, it is evaluated at the original in-context example count $N$ rather than the number of compressed slots $M$. Consequently, $\sum_j w_j=N$. For any query batch $Q$, the attention-output matching loss is
\begin{equation}
    \mathcal L_{\mathrm{op}}(Q)
    =\left\|P(Q;\widetilde K,w)\widetilde V-\fullop(Q)\right\|_F^2.
    \label{eq:operator-compression-target}
\end{equation}
Here $Q$ denotes an arbitrary query batch; compilation uses the in-context-example-derived $Q_{\mathrm{cal}}$ defined below and no test-set information.

% End sections/formulation

% Begin sections/method
\subsection{Joint KV Prototype Construction}
\label{sec:prototype-construction}

Within each head, we run deterministic Lloyd K-means on concatenated states $z_i=[k_i/\sigma_K;v_i/\sigma_V]\in\Real^{2d}$, where $\sigma_K=\operatorname{Std}(K)$ and $\sigma_V=\operatorname{Std}(V)$ are scalar standard deviations over all matrix entries. Empty clusters are repaired by assigning farthest points. Let nonempty $C_j\subseteq\{1,\ldots,N\}$ denote the final Lloyd assignment used to compute prototype $j\in\{1,\ldots,M\}$. Joint clustering avoids merging states with similar keys but incompatible values. The initial compressed states are
\begin{equation}
    \widetilde k_j=\frac{1}{|C_j|}\sum_{i\in C_j}k_i,
    \quad
    \widetilde v_{0j}=\frac{1}{|C_j|}\sum_{i\in C_j}v_i.
    \label{eq:initial-prototypes}
\end{equation}
$\widetilde V_0$ denotes the initial value vectors obtained from the cluster means. After constructing the joint-KV prototypes, we reassign all in-context examples to their nearest prototypes in the same normalized space and define the cluster count as $w_j=|D_j|$, where $D_j$ is the resulting group. Thus $w_j>0$ and $\sum_j w_j=N$. Here, $C_j$ determines the prototype means, whereas $D_j$ determines the exact count associated with each final prototype; empty-group handling and numerical settings are given in the appendix. Representing the $w_j$ keys in $D_j$ by $\widetilde k_j$ gives the combined softmax numerator
\begin{equation}
    \sum_{i\in D_j}\exp\langle\phi_N(q),\widetilde k_j\rangle
    =w_j\exp\langle\phi_N(q),\widetilde k_j\rangle
    =\exp\!\left(\langle\phi_N(q),\widetilde k_j\rangle+\log w_j\right).
    \label{eq:multiplicity-identity}
\end{equation}
Thus $\log w_j$ accounts for the number of source states represented by each compressed slot. It does not remove the approximation error between $k_i$ and $\widetilde k_j$; $\phi_N$ separately keeps the backbone's query transformation unchanged, including any source-length dependence.

\subsection{Constructing Calibration Equations}

Fitting the compressed value vectors requires query--output pairs for the attention layer. During the pass that constructs the full cache, the model produces one attention query vector per in-context example at every layer and head. We use all $N$ vectors, so $N_{\mathrm{cal}}=N$. For a fixed layer and head, stack them as
\begin{equation}
\begin{aligned}
    Q_{\mathrm{cal}}
    &=[(q_1^{\mathrm{cal}})^\top;\ldots;(q_{N_{\mathrm{cal}}}^{\mathrm{cal}})^\top]
      \in\Real^{N_{\mathrm{cal}}\times d},\\
    \Phi_N(Q_{\mathrm{cal}})
    &=[\phi_N(q_1^{\mathrm{cal}})^\top;\ldots;\phi_N(q_{N_{\mathrm{cal}}}^{\mathrm{cal}})^\top]
      \in\Real^{N_{\mathrm{cal}}\times d}.
\end{aligned}
    \label{eq:calibration-queries}
\end{equation}
Here $N$ is both the original in-context example count used by $\phi_N$ and the number of fitting states. No external calibration set or held-out query rows are required.

Applying the full cache to $Q_{\mathrm{cal}}$ gives the target outputs
\begin{equation}
    Y_{\mathrm{cal}}=\fullop(Q_{\mathrm{cal}})
    =\softmax_{\mathrm{row}}(\Phi_N(Q_{\mathrm{cal}})K^\top)V,
    \label{eq:reference-output}
\end{equation}
while the compressed keys and cluster sizes give the corresponding attention weights
\begin{equation}
    P_{\mathrm{cal}}=P(Q_{\mathrm{cal}};\widetilde K,w)\in\Real^{N_{\mathrm{cal}}\times M}.
\end{equation}
With the compressed keys and cluster-count weights fixed, value fitting therefore reduces to the matrix regression problem
\begin{equation}
    P_{\mathrm{cal}}\widetilde V \approx Y_{\mathrm{cal}}.
    \label{eq:calibration-equation}
\end{equation}
\subsection{Value Fitting with Fixed Attention Weights}
\label{sec:value-fitting}

Keys determine the attention weights, whereas values determine the content returned under those weights. Jointly fitting both would require
\begin{equation*}
    \min_{\widetilde K,\widetilde V}
    \left\|\softmax_{\mathrm{row}}\!\left(\Phi_N(Q_{\mathrm{cal}})\widetilde K^\top+
    \mathbf 1_{N_{\mathrm{cal}}}(\log w)^\top\right)\widetilde V-Y_{\mathrm{cal}}\right\|_F^2,
\end{equation*}
a non-convex problem because changing $\widetilde K$ also changes the attention weights. With $(\widetilde K,w)$ fixed, $P_{\mathrm{cal}}$ is fixed and the output $P_{\mathrm{cal}}\widetilde V$ is linear in $\widetilde V$. We therefore keep the clustered keys fixed and fit only the returned values.

For each attention head in every layer, compute its regularization scale $\gamma=\gamma(P_{\mathrm{cal}})$ as
\begin{equation}
    \gamma(P_{\mathrm{cal}})=\max\!\left\{\frac{1}{M}\operatorname{tr}(P_{\mathrm{cal}}^\top P_{\mathrm{cal}}),\epsilon_{\mathrm{num}}\right\}.
    \label{eq:ridge-scale}
\end{equation}
Here $\epsilon_{\mathrm{num}}>0$ prevents a degenerate zero scale.
\method{} optimizes only values through scale-normalized anchored ridge regression:
\begin{equation}
    \widetilde V^*
    =\argmin_{Z\in\Real^{M\times d}}
    \|P_{\mathrm{cal}}Z-Y_{\mathrm{cal}}\|_F^2
    +\lambda\gamma(P_{\mathrm{cal}})\|Z-\widetilde V_0\|_F^2.
    \label{eq:ridge-objective}
\end{equation}
The unique solution is
\begin{equation}
    \boxed{\;
    \widetilde V^*=(P_{\mathrm{cal}}^\top P_{\mathrm{cal}}+\lambda\gamma I_M)^{-1}
    (P_{\mathrm{cal}}^\top Y_{\mathrm{cal}}+\lambda\gamma\widetilde V_0)
    \;}.
    \label{eq:closed-form}
\end{equation}
$P_{\mathrm{cal}}^\top P_{\mathrm{cal}}$ is the Gram matrix of the least-squares design. When the numerical floor is inactive, $\gamma=\|P_{\mathrm{cal}}\|_F^2/M$, its mean diagonal value. Scaling the anchor by $\gamma$ keeps the relative regularization strength comparable across attention heads, while stabilizing directions weakly constrained by the fitting states.

% End sections/method

% Begin sections/theory
\subsection{Design Analysis}
\label{sec:theory}

We first characterize the error introduced by replacing groups of in-context-example states with weighted joint-KV prototypes. Assume every $D_j$ is nonempty, $\|v_i\|_2\leq V_{\max}$, $w_j=|D_j|$, and, for every $i\in D_j$,
$|\langle\phi_N(q),k_i-\widetilde k_j\rangle|\leq\epsilon$ and
$\|v_i-\widetilde v_{0j}\|_2\leq\delta$. Grouping the original attention terms then gives
\begin{equation}
 \|\fullop(q)-\widehat{\mathcal A}_{\mathcal D,0}(q)\|_2
 \leq \delta+2V_{\max}\tanh(\epsilon).
 \label{eq:cluster-bound}
\end{equation}
Thus, the approximation improves as keys and values become more concentrated within each cluster. Exact cluster counts remain necessary even at zero cluster radius: identical members of cluster $j$ contribute the factor $|D_j|\exp(\langle\phi_N(q),\widetilde k_j\rangle)$ before softmax normalization, so removing $\log w_j$ changes the attention weights. Likewise, a length-aware model must continue to evaluate $\phi_N$ using the original example count $N$, rather than the retained count $M$.

Value fitting provides a complementary guarantee. For fixed $(\widetilde K,w)$ and $\lambda\gamma(P_{\mathrm{cal}})>0$, the objective in \Eqref{eq:ridge-objective} is strictly convex and has the unique solution in \Eqref{eq:closed-form}. Since the initialization $\widetilde V_0$ is feasible, optimality implies
\begin{equation}
 \|P_{\mathrm{cal}}\widetilde V^*-Y_{\mathrm{cal}}\|_F^2
 \leq
 \|P_{\mathrm{cal}}\widetilde V_0-Y_{\mathrm{cal}}\|_F^2.
 \label{eq:training-monotonicity}
\end{equation}
The guarantee applies only to the calibration queries with fixed attention weights: it does not ensure lower test error or higher predictive accuracy. This distinction matches the empirical pattern in Figure~\ref{fig:evidence}: value fitting consistently improves attention-output fidelity, while its effect on accuracy remains dataset dependent. Proofs and additional analysis appear in Appendix~\ref{app:proofs}.

% End sections/theory

% Begin sections/experiments
\section{Experiments}
\label{sec:experiments}

We evaluate \method{} in terms of predictive accuracy, long-context scalability, component effects, and multi-query serving efficiency. We first compare it with compression and retrieval methods at the same retained counts, and then use long-table, component, and serving experiments to examine its accuracy retention, the role of each design choice, and the efficiency gains from reuse across queries.

\subsection{Experimental Protocol}

\paragraph{Primary and long-context benchmarks.}
The primary benchmark contains 64 held-out OpenML-CC18 datasets~\citep{openmlcc18}, uses up to 512 in-context examples, and evaluates retained counts $M\in\{32,64\}$; ten disjoint validation datasets determine all hyperparameters. Long-context experiments cover seven CC18 datasets together with SUSY, HIGGS, and Covertype under varying in-context example counts and compression ratios. Appendix~\ref{app:details} gives further protocol and data-split details.

\paragraph{Baselines.}
We compare full-context inference, fixed row selection, dynamic retrieval, and direct KV compression. Fixed row selection includes class-balanced random selection, input-space K-means, and LUCoS~\citep{lucos}; dynamic retrieval includes MICP~\citep{mixturepfn}, CRUMB~\citep{crumb}, KATE~\citep{kate}, standardized raw-space KNN, and an ARASH-style reproduction~\citep{arash}. The primary direct-KV baseline is weighted joint-KV; key centroids and ToMe/KVMerger-style variants are compared in Appendix Table~\ref{tab:matched-compression}. All methods use the same prediction model; Appendix~\ref{app:baselines} gives exact definitions.

\paragraph{Metrics and uncertainty.}
We use accuracy averaged equally across datasets to measure predictive performance; prediction agreement, probability mean absolute error (MAE), and attention-output mean squared error (MSE) measure deviation from full-context inference. We also report cache memory and serving time. Confidence intervals use paired bootstrap over datasets; Appendix~\ref{app:details} defines the metrics and Appendix~\ref{app:scale-matrix} gives the timing protocol.

\subsection{Accuracy at the Same Retained Count}

We first compare methods at the same retained count $M$: row-selection methods retain $M$ in-context examples per prediction, while cache-compression methods retain $M$ KV slots per layer and head. Because $M$ in-context examples likewise produce $M$ KV slots per layer and head, the two families have the same online attention length. \method{} additionally stores $M$ log cluster counts $\log w_j$ and the scalar original example count $N$; this metadata is negligible relative to the $M$ $d$-dimensional key and value vectors. Table~\ref{tab:main} reports results on all 64 held-out datasets.

\begin{table}[t]
\centering
\caption{\textbf{Held-out CC18 classification.} Mean accuracy across 64 datasets at retained counts $M=32$ and $M=64$. Full uses all available in-context examples; bold marks the best compression or retrieval method. Appendix Table~\ref{tab:matched-compression} reports additional direct-compression methods.}
\label{tab:main}
\small
\begin{tabular}{lcc}
\toprule
Method & $M=32$ & $M=64$ \\
\midrule
\multicolumn{3}{l}{\emph{Global row selection}} \\
\quad Random & 0.6819 & 0.7404 \\
\quad Input K-means & 0.7321 & 0.7925 \\
\quad LUCoS & 0.7348 & 0.7922 \\
\addlinespace[2pt]
\multicolumn{3}{l}{\emph{Dynamic row selection}} \\
\quad MICP & 0.7905 & 0.8137 \\
\quad CRUMB & 0.8033 & 0.8296 \\
\quad KATE & 0.8088 & 0.8344 \\
\quad Standardized raw-space KNN & 0.8117 & 0.8355 \\
\quad ARASH-style & 0.7910 & 0.8145 \\
\addlinespace[2pt]
\multicolumn{3}{l}{\emph{KV compression}} \\
\quad Weighted joint-KV & 0.8391 & 0.8450 \\
\quad \textbf{\method{}} & \textbf{0.8534} & \textbf{0.8566} \\
\midrule
Full context & 0.8617 & 0.8617 \\
\bottomrule
\end{tabular}
\vspace{-7pt}
\end{table}

At the same online attention length, \method{} gives the strongest average accuracy among the compared reusable and query-specific alternatives (Table~\ref{tab:main}). Its advantage over weighted joint-KV indicates that compression quality is not explained by clustering alone: matching the full-cache responses contributes beyond retaining cluster mass. Increasing $M$ narrows the gap to Full, as a larger memory recovers some accuracy lost to compression. The paired intervals in Appendix Table~\ref{tab:main-pairwise} are positive against the strongest retrieval and direct-compression baselines. Moving from $M=32$ to $64$ raises the best retrieval accuracy by 2.38 percentage points, versus 0.32 points for \method{}, suggesting that retrieval benefits more from the extra rows.

\subsection{Scaling with In-Context Example Count and Retained Count}

We then evaluate \method{} at different in-context example counts $N$, with the retained count set to $M=N/32$, $N/16$, and $N/8$, to examine its performance across context lengths and compression ratios. Table~\ref{tab:scale-matrix} reports all 12 settings. Every setting uses the same seven datasets, so differences reflect the operating regime rather than a changing benchmark composition. Appendix~\ref{app:baselines} specifies the implementation and scope of each baseline.

\begin{table*}[t]
\centering
\caption{\textbf{Accuracy across in-context example counts and retained counts.} Means over seven CC18 datasets. Within each in-context example count block, all compressed methods use the same retained count $M=N/32,N/16,N/8$. Bold marks the highest mean accuracy among compressed methods. $\Delta$ is QCOC minus the best other compressed method, including global selection, dynamic selection, and KV compression but excluding Full, in percentage points (pp); positive values favor QCOC.}
\label{tab:scale-matrix}
\scriptsize
\setlength{\tabcolsep}{2.7pt}
\resizebox{\textwidth}{!}{%
\begin{tabular}{lccc@{\hspace{5pt}}ccc@{\hspace{5pt}}ccc@{\hspace{5pt}}ccc}
\toprule
& \multicolumn{3}{c}{$N=512$} & \multicolumn{3}{c}{$N=2{,}048$} & \multicolumn{3}{c}{$N=4{,}096$} & \multicolumn{3}{c}{$N=8{,}192$} \\
\cmidrule(lr){2-4}\cmidrule(lr){5-7}\cmidrule(lr){8-10}\cmidrule(lr){11-13}
Method & 16 & 32 & 64 & 64 & 128 & 256 & 128 & 256 & 512 & 256 & 512 & 1,024 \\
\midrule
\multicolumn{13}{l}{\emph{Global row selection}} \\
\quad Random & .6003 & .6276 & .6763 & .6716 & .7120 & .7254 & .6854 & .7421 & .7494 & .7319 & .7487 & .7596 \\
\quad Input K-means & .6656 & .6846 & .7370 & .7153 & .7440 & .7573 & .7389 & .7610 & .7759 & .7399 & .7793 & .7871 \\
\quad LUCoS & .6463 & .6721 & .7323 & .7067 & .7266 & .7696 & .7374 & .7626 & .7773 & .7656 & .7670 & .7900 \\
\addlinespace[2pt]
\multicolumn{13}{l}{\emph{Dynamic row selection}} \\
\quad MICP & .6549 & .7017 & .7016 & .7483 & .7691 & .7900 & .7706 & .7853 & .7881 & .7844 & .7930 & .7923 \\
\quad CRUMB & .6739 & .7057 & .7417 & .7557 & .7756 & .7899 & .7746 & .7933 & .7956 & .7909 & .7976 & .8061 \\
\quad KATE & .6960 & .7233 & .7456 & .7584 & .7771 & .7903 & .7883 & .7963 & \textbf{.8039} & .7997 & .8019 & \textbf{.8083} \\
\quad Standardized raw-space KNN & .7007 & .7246 & .7411 & .7696 & .7843 & .7939 & .7759 & .7924 & .7976 & .7976 & .8051 & .8064 \\
\quad ARASH-style & .6966 & .6734 & .6930 & .7411 & .7624 & .7639 & .7650 & .7764 & .7773 & .7760 & .7851 & .7827 \\
\addlinespace[2pt]
\multicolumn{13}{l}{\emph{KV compression}} \\
\quad Weighted joint-KV & .7634 & .7681 & .7723 & .7861 & .7896 & .7916 & .7903 & .7904 & .7939 & .7946 & .7979 & .7997 \\
\quad \method{} & \textbf{.7663} & \textbf{.7763} & \textbf{.7769} & \textbf{.7934} & \textbf{.7953} & \textbf{.8000} & \textbf{.8026} & \textbf{.8024} & .8033 & \textbf{.8056} & \textbf{.8081} & .8069 \\
\quad $\Delta$ vs. best other (pp) & $+0.29$ & $+0.81$ & $+0.46$ & $+0.73$ & $+0.57$ & $+0.61$ & $+1.23$ & $+0.61$ & $-0.06$ & $+0.59$ & $+0.30$ & $-0.14$ \\
\midrule
Full & .7784 & .7784 & .7784 & .8000 & .8000 & .8000 & .8020 & .8020 & .8020 & .8077 & .8077 & .8077 \\
\bottomrule
\end{tabular}}
\end{table*}

Table~\ref{tab:scale-matrix} reveals a regime change rather than a uniform margin. \method{} leads in 10 of 12 settings and is most clearly favored when the retained count is small relative to the available context. As the budget is relaxed, query-specific retrieval becomes competitive and narrowly leads only at the largest $M$ in the $N=4{,}096$ and $8{,}192$ blocks. Across the full grid, however, \method{} remains only 0.23 percentage points below Full on average. The main benefit of compiling a shared operator is therefore robustness under tight compression; the reversals at larger budgets suggest that local retrieval benefits more from additional retained examples, not merely from a longer source context.

\enlargethispage{0.6\baselineskip}
\subsection{Component Analysis}

The factorial analysis in Appendix Table~\ref{tab:components} identifies two distinct requirements. Cluster-count weighting and preserving the original count $N$ must be used together: either correction alone leaves the compressed attention computation inconsistent, and their effects are strongly non-additive. Once these structural terms are fixed, fitting only $\widetilde V$ supplies the remaining accuracy gain. Thus, the count terms correct attention weighting and query transformation, while value fitting calibrates the resulting output; these roles are complementary. Joint-KV and key-only prototypes have statistically indistinguishable final accuracy, but joint-KV yields more consistently lower reconstruction and prediction errors; its main contribution is therefore fidelity, not a separately identifiable accuracy gain.

\begin{figure}[H]
    \centering
    \includegraphics[width=0.98\textwidth]{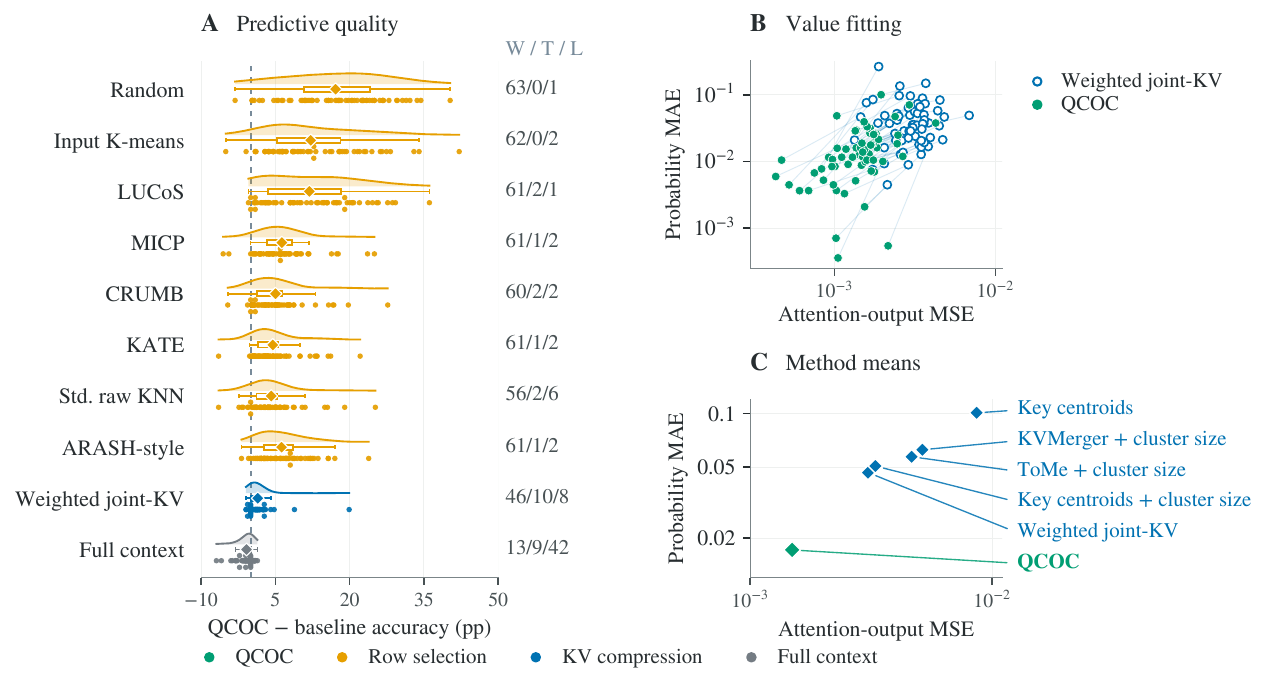}
    \caption{\textbf{Accuracy and fidelity at $M=32$ on 64 CC18 datasets.} \textbf{A:} QCOC accuracy minus every Table~\ref{tab:main} baseline on the full test split. Dots are datasets, diamonds means, and boxes interquartile ranges, not confidence intervals. \textbf{B/C:} B shows errors before (open) and after (filled) value fitting; C shows mean errors for Appendix Table~\ref{tab:matched-compression} methods. B and C use logarithmic axes.}
    \label{fig:evidence}
\end{figure}

Figure~\ref{fig:evidence} separates that mechanistic claim from the downstream task claim. The paired accuracy distribution in panel A is heterogeneous: gains over weaker row-selection baselines are broad, while comparisons with stronger retrieval and KV baselines are tighter. In contrast, every dataset moves toward Full after value fitting in panel B, and panel C shows that this advantage remains when other direct-compression constructions are included. Panels B/C thus test the proposed operator approximation more directly than classification accuracy alone. Value fitting reliably improves full-cache fidelity; whether that changes the predicted class still depends on each dataset's decision margins.

\subsection{Long-Context Accuracy and Serving Cost}

The long-table results compare accuracy, cache footprint, and online serving cost. At $N=8{,}192$ and $M=512$, \method{} is the strongest compressed method on CC18-Long, SUSY, and HIGGS, whereas KATE remains substantially better on Covertype. Relative to Full, its cache is 10.53$\times$ smaller on CC18-Long; Appendix Table~\ref{tab:long-external} gives the corresponding results for the three external datasets. Increasing the retained count narrows but does not close the Covertype gap to Full (Appendix Tables~\ref{tab:external-splits} and~\ref{tab:covertype-adaptive}). This failure case matters: one shared memory is effective when the full-cache behavior can be represented globally, but query-local retrieval can be preferable on heterogeneous tables.

\begin{figure}[t]
\centering
\includegraphics[width=0.90\textwidth]{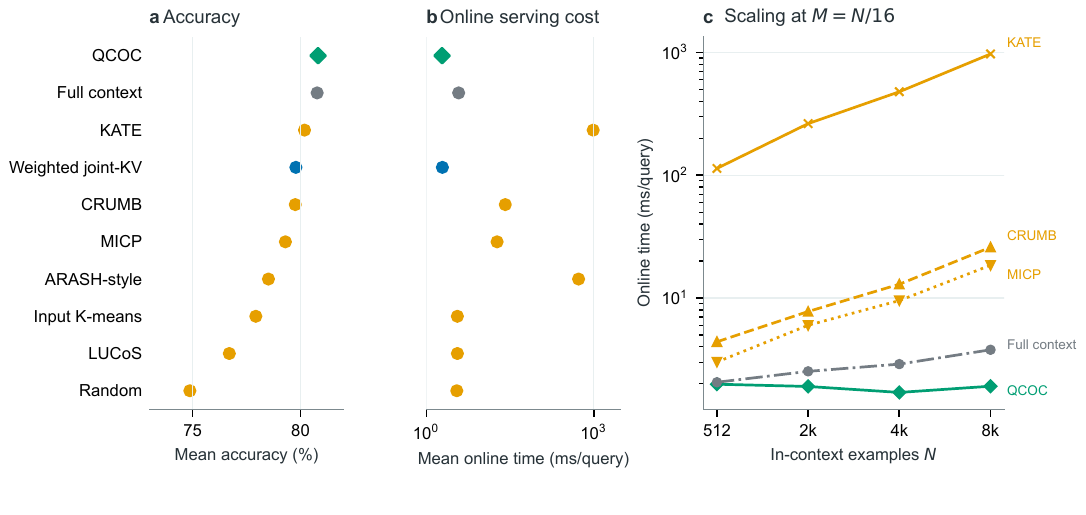}
\caption{\textbf{Accuracy and online serving cost.} \textbf{A/B:} Accuracy and matched single-core serving time at $N=8{,}192$, $M=512$. \textbf{C:} Serving-time scaling at $M=N/16$. Times average 1,000-query batches; B/C use logarithmic axes. One-time setup and \method{} compilation are excluded.}
\label{fig:efficiency-tradeoff}
\end{figure}

After compilation, the shared memory reduces single-core serving time from 3.79 to 1.91 ms per query relative to Full and avoids KATE's per-query cache construction (Figure~\ref{fig:efficiency-tradeoff}). In panel C, \method{} time changes little as $N$ grows at a fixed compression ratio, while Full and query-specific retrieval become costlier. The advantage comes from reusing a bounded cache, not an inherent speed difference between KV-compression constructions. These measurements exclude one-time cache construction and compilation; Appendix Table~\ref{tab:compile-time} reports the latter separately. Hence, practical gains require enough queries to amortize compilation. The positive residual--accuracy-loss association in Appendix Figure~\ref{fig:residual-diagnostic} motivates reconstruction residual as a candidate diagnostic.

\FloatBarrier

% End sections/experiments

% Begin sections/discussion
\section{Conclusion}
\label{sec:conclusion}

This work treats compression for tabular in-context learning as compilation of the attention operator induced by a fixed example set, rather than selection of a smaller set of rows. \method{} approximates that operator with a compact, reusable KV memory and calibrates its values to preserve query-dependent attention outputs. On held-out CC18 datasets, it achieves the highest mean accuracy among the compared compression and retrieval methods at both retained counts; on long tables, it remains close to Full in aggregate while reducing online serving cost across repeated queries.

These gains depend on reusing the same compiled memory across enough queries to justify full-cache construction and one-time compilation. A single shared memory can also miss query-local structure, as the Covertype result shows. Reducing compilation cost and selecting among multiple reusable memories are therefore natural next steps.

% End sections/discussion

\bibliographystyle{cite}
\bibliography{references}

\appendix

% Begin sections/appendix
\section{Supplementary Design Analysis}

For interpretation at network level, let $\varepsilon_\ell(q_\ell)$ bound the complete block perturbation at a common input, including all heads, output projection, and subsequent within-block operations; let $\rho_r$ bound the complete downstream block's Lipschitz constant. For $L$ compressed blocks, a hybrid-network argument gives
\begin{equation}
 \|f_{\mathcal D}(x)-\widehat f_{\mathcal D}(x)\|_2
 \leq \sum_{\ell=1}^{L}\left(\prod_{r=\ell+1}^{L}\rho_r\right)\varepsilon_\ell(q_\ell).
 \label{eq:propagation}
\end{equation}
A $\rho_g$-Lipschitz prediction head multiplies the right side by $\rho_g$.

Differences between fitting and deployment queries provide another source of error. Define $\ell_{\widetilde V}(q)=\|\compop(q)-\fullop(q)\|_2^2$, and let $\mu_{\mathrm{cal}}$ and $\mu_{\mathrm{test}}$ denote their distributions under metric $d_q$. If $\ell_{\widetilde V}$ is $L_q$-Lipschitz, then
\begin{equation}
 \mathbb E_{\mu_{\mathrm{test}}}\ell_{\widetilde V}
 \leq \mathbb E_{\mu_{\mathrm{cal}}}\ell_{\widetilde V}
 +L_qW_1(\mu_{\mathrm{cal}},\mu_{\mathrm{test}}),
 \label{eq:shift}
\end{equation}
where $W_1$ is the 1-Wasserstein distance under $d_q$. These bounds clarify possible error accumulation and distribution shift; they are not deployment-performance guarantees, and the experiments do not estimate their Lipschitz or Wasserstein terms.

\section{Derivations}
\label{app:proofs}

\subsection{Derivation of the grouped approximation bound}
Fix a query $q$ and abbreviate
\begin{equation}
 s_i=\langle\phi_N(q),k_i\rangle,
 \qquad
 \widetilde s_j=\langle\phi_N(q),\widetilde k_j\rangle.
\end{equation}
For group $D_j$, define its compressed and exact unnormalized weights by
\begin{equation}
 a_j=|D_j|e^{\widetilde s_j},
 \qquad
 b_j=\sum_{i\in D_j}\exp(\langle\phi_N(q),k_i\rangle).
\end{equation}
For every $i\in D_j$, the logit-radius assumption implies
$e^{-\epsilon}e^{\widetilde s_j}\leq e^{s_i}\leq
e^{\epsilon}e^{\widetilde s_j}$. Summing these inequalities over $D_j$ gives
\begin{equation}
 e^{-\epsilon}a_j\leq b_j\leq e^\epsilon a_j.
 \label{eq:proof-group-mass}
\end{equation}
Let $A=\sum_j a_j$, $B=\sum_j b_j$, and let
$\alpha_j=a_j/A$ and $\beta_j=b_j/B$ be the compressed and exact
probabilities assigned to group $j$. Summing
\Eqref{eq:proof-group-mass} over groups also gives
$e^{-\epsilon}A\leq B\leq e^\epsilon A$. Hence
\begin{equation}
 e^{-2\epsilon}
 \leq \frac{\beta_j}{\alpha_j}
 =\frac{b_j/a_j}{B/A}
 \leq e^{2\epsilon}.
 \label{eq:proof-likelihood-ratio}
\end{equation}
For any $r\in[e^{-2\epsilon},e^{2\epsilon}]$,
\begin{equation}
 |r-1|\leq \frac{e^{2\epsilon}-1}{e^{2\epsilon}+1}(r+1)
 =\tanh(\epsilon)(r+1).
\end{equation}
Applying this inequality to $r_j=\beta_j/\alpha_j$ and using
$\sum_j\alpha_j=\sum_j\beta_j=1$ yields
\begin{equation}
 \|\beta-\alpha\|_1
 =\sum_j\alpha_j|r_j-1|
 \leq 2\tanh(\epsilon).
 \label{eq:proof-group-tv}
\end{equation}

It remains to control the values returned within each group. Define the exact
conditional weights and conditional value mean
\begin{equation}
 \pi_{i\mid j}=\frac{e^{s_i}}{b_j},
 \qquad
 \bar v_j(q)=\sum_{i\in D_j}\pi_{i\mid j}v_i.
\end{equation}
Because the $\pi_{i\mid j}$ are nonnegative and sum to one,
\begin{align}
 \|\bar v_j(q)-\widetilde v_{0j}\|_2
 &\leq \sum_{i\in D_j}\pi_{i\mid j}
       \|v_i-\widetilde v_{0j}\|_2
 \leq\delta, \\
 \|\bar v_j(q)\|_2
 &\leq \sum_{i\in D_j}\pi_{i\mid j}\|v_i\|_2
 \leq V_{\max}.
\end{align}
The exact and compressed outputs can now be written as
$\fullop(q)=\sum_j\beta_j\bar v_j(q)$ and
$\widehat{\mathcal A}_{\mathcal D,0}(q)=\sum_j\alpha_j\widetilde v_{0j}$.
Adding and subtracting $\sum_j\alpha_j\bar v_j(q)$ gives
\begin{align}
 \|\fullop(q)-\widehat{\mathcal A}_{\mathcal D,0}(q)\|_2
 &\leq
 \left\|\sum_j(\beta_j-\alpha_j)\bar v_j(q)\right\|_2
 +\left\|\sum_j\alpha_j(\bar v_j(q)-\widetilde v_{0j})\right\|_2\\
 &\leq V_{\max}\|\beta-\alpha\|_1+\delta
 \leq 2V_{\max}\tanh(\epsilon)+\delta,
\end{align}
which is \Eqref{eq:cluster-bound}. \hfill$\square$

\subsection{Derivation of the closed-form value fit}
Write $P=P_{\mathrm{cal}}$, $Y=Y_{\mathrm{cal}}$, and
$\eta=\lambda\gamma(P)>0$. For
$F(Z)=\|PZ-Y\|_F^2+\eta\|Z-\widetilde V_0\|_F^2$, the second directional
derivative along any nonzero $H\in\Real^{M\times d}$ is
\begin{equation}
 \frac{d^2}{dt^2}F(Z+tH)\bigg|_{t=0}
 =2\|PH\|_F^2+2\eta\|H\|_F^2>0.
\end{equation}
Thus $F$ is strictly convex and has at most one minimizer. Its gradient is
\begin{equation}
 \nabla F(Z)=2P^\top(PZ-Y)+2\eta(Z-\widetilde V_0).
\end{equation}
Setting the gradient to zero gives the normal equation
\begin{equation}
 (P^\top P+\eta I)Z=P^\top Y+\eta\widetilde V_0.
 \label{eq:proof-normal-equation}
\end{equation}
For every nonzero $u\in\Real^M$,
$u^\top(P^\top P+\eta I)u=\|Pu\|_2^2+\eta\|u\|_2^2>0$.
The coefficient matrix in \Eqref{eq:proof-normal-equation} is therefore
positive definite and invertible. Solving the equation gives
\Eqref{eq:closed-form}; strict convexity makes this stationary point the unique
global minimizer.

Finally, optimality relative to the feasible anchor $\widetilde V_0$ gives
\begin{align}
 &\|P\widetilde V^*-Y\|_F^2
 +\eta\|\widetilde V^*-\widetilde V_0\|_F^2\\
 &\qquad\leq
 \|P\widetilde V_0-Y\|_F^2
 +\eta\|\widetilde V_0-\widetilde V_0\|_F^2
 =\|P\widetilde V_0-Y\|_F^2,
\end{align}
Dropping the nonnegative anchor term on the left proves
\Eqref{eq:training-monotonicity}.
\hfill$\square$

\subsection{Derivation of the layer-by-layer bound}
To connect single-head errors to block errors, fix a common block input $h$ and let $e_{\ell,a}(h)$ bound the Euclidean output error of head $a$ before concatenation. Let $W^O_\ell$ be the attention output projection and let $c_\ell$ bound the Lipschitz constant of the remaining within-block map with respect to its projected attention input, holding the residual input fixed. Then a valid local bound is
\begin{equation}
 \varepsilon_\ell(q_\ell(h))
 =c_\ell\|W^O_\ell\|_2
   \left(\sum_{a=1}^{H_\ell}e_{\ell,a}(h)^2\right)^{1/2}.
 \label{eq:head-to-block}
\end{equation}
Concatenation gives the root-sum-of-squares factor; the shared output bias and attention residual cancel at a common input. The factor $c_\ell$ includes subsequent normalization and feed-forward/residual operations; for a remaining map $u\mapsto u+\operatorname{FFN}(\operatorname{Norm}(u))$, it is at most $1+L_{\mathrm{FFN}}L_{\mathrm{Norm}}$. Query preprocessing is included in the head errors evaluated at $h$, and its sensitivity to differing inputs is included in the complete-block constants $\rho_r$. Thus a single-head calibration MSE is not itself $\varepsilon_\ell$ or a uniform bound on deployment states.

Let $G_\ell$ and $\widehat G_\ell$ denote the exact and compressed maps at
block $\ell$, including the common residual and feed-forward operations, and
let the local perturbation satisfy
\begin{equation}
 \|G_\ell(h)-\widehat G_\ell(h)\|_2
 \leq\varepsilon_\ell(q_\ell)
\end{equation}
at every intermediate state encountered below. Define the hybrid output
\begin{equation}
 H^{(\ell)}=
 G_L\circ\cdots\circ G_{\ell+1}\circ
 \widehat G_\ell\circ\cdots\circ\widehat G_1(x),
 \qquad \ell=0,\ldots,L,
\end{equation}
where $H^{(0)}=f_{\mathcal D}(x)$ and
$H^{(L)}=\widehat f_{\mathcal D}(x)$. Adjacent hybrids share the same
compressed prefix through block $\ell-1$ and differ only in whether block
$\ell$ is exact or compressed. If $h_{\ell-1}$ is the output of that common
prefix, Lipschitz continuity of the exact downstream blocks gives
\begin{align}
 \|H^{(\ell)}-H^{(\ell-1)}\|_2
 &\leq
 \left(\prod_{r=\ell+1}^{L}\rho_r\right)
 \|\widehat G_\ell(h_{\ell-1})-G_\ell(h_{\ell-1})\|_2\\
 &\leq
 \left(\prod_{r=\ell+1}^{L}\rho_r\right)
 \varepsilon_\ell(q_\ell).
\end{align}
Using the telescoping identity
$H^{(L)}-H^{(0)}=\sum_{\ell=1}^L(H^{(\ell)}-H^{(\ell-1)})$
and the triangle inequality proves \Eqref{eq:propagation}. If a prediction
head $g$ follows the last block, then
$\|g(H^{(L)})-g(H^{(0)})\|_2\leq\rho_g\|H^{(L)}-H^{(0)}\|_2$,
which gives the final statement. \hfill$\square$

\subsection{Derivation of the query-distribution bound}
Let $\Pi(\mu_{\mathrm{cal}},\mu_{\mathrm{test}})$ denote the set of
couplings of the two distributions, and fix any
$\pi\in\Pi(\mu_{\mathrm{cal}},\mu_{\mathrm{test}})$. Because the marginals
of $\pi$ are $\mu_{\mathrm{cal}}$ and $\mu_{\mathrm{test}}$,
\begin{align}
\mathbb E_{q'\sim\mu_{\mathrm{test}}}\ell_{\widetilde V}(q')
-\mathbb E_{q\sim\mu_{\mathrm{cal}}}\ell_{\widetilde V}(q)
&=\mathbb E_{(q,q')\sim\pi}\!\left[\ell_{\widetilde V}(q')-\ell_{\widetilde V}(q)\right]\\
&\leq \mathbb E_{(q,q')\sim\pi}
 \left|\ell_{\widetilde V}(q')-\ell_{\widetilde V}(q)\right|\\
&\leq L_q\,\mathbb E_{(q,q')\sim\pi}d_q(q,q').
\end{align}
The left side is independent of the chosen coupling. Taking the infimum of
the right side over $\pi\in\Pi(\mu_{\mathrm{cal}},\mu_{\mathrm{test}})$ and
using the definition of the 1-Wasserstein distance gives
\begin{equation}
 \mathbb E_{\mu_{\mathrm{test}}}\ell_{\widetilde V}
 -\mathbb E_{\mu_{\mathrm{cal}}}\ell_{\widetilde V}
 \leq L_q W_1(\mu_{\mathrm{cal}},\mu_{\mathrm{test}}),
\end{equation}
which rearranges to \Eqref{eq:shift}. \hfill$\square$

\section{Additional Experimental Details}
\label{app:details}

\paragraph{Data and splits.}
For the primary study, we retain 64 OpenML-CC18 tasks with at most ten classes after excluding the flattened-image tables. A stratified 20\% test partition is held out, and up to 512 in-context examples are selected from the remainder. Ten disjoint validation datasets determine all hyperparameters. Cache compression receives no test query, label, prediction, or test-query hidden state. We use a single-estimator TabICL v2 classifier with column and class shuffling disabled.

The seven CC18 long-table datasets use $N\in\{512,2048,4096,8192\}$, $M\in\{N/32,N/16,N/8\}$, and 1,000 test queries per setting. We use chronology-respecting holdouts for \texttt{electricity} and \texttt{bank-marketing} and stratified holdouts otherwise. External experiments use official Covertype, SUSY, and HIGGS data with $N=8192$, $M=512$, 1,000 disjoint test rows, and three splits. Dataset identifiers are listed below.

\paragraph{Baseline execution.}
MICP routes queries among cluster-specific contexts; CRUMB constructs an MMD-matched context for each query group; KATE performs per-query retrieval in TabClustPFN embedding space; and standardized raw-space KNN retrieves directly from standardized inputs. Every baseline uses TabICL for prediction rather than label voting, and all methods share the same model and in-context example pool.

\paragraph{Statistical unit.}
The dataset is the unit for paired bootstrap intervals. For repeated-split diagnostics, scores are first averaged within each dataset before datasets are resampled, so repeated splits do not count as independent evidence.

\paragraph{Metric definitions.}
For a regression test split with $T$ targets, define $\bar y=T^{-1}\sum_{t=1}^T y_t$ and
\begin{equation}
 \operatorname{nRMSE}
 =\frac{\sqrt{T^{-1}\sum_{t=1}^T(\widehat y_t-y_t)^2}}
 {\max\{\sqrt{T^{-1}\sum_{t=1}^T(y_t-\bar y)^2},10^{-12}\}}.
\end{equation}
TabArena results average the three official splits within each dataset and then average the 13 datasets equally.

For one dataset and split, let $E_{\ell,\mathrm{fit}}(M)$ and $E_{\ell,0}(M)$ be attention-output MSEs after value fitting and at the weighted joint-KV anchor, respectively, on the same fixed attention query vectors produced by the in-context examples. Each MSE averages all query, head, and value-coordinate elements within layer $\ell$, before output projection. We define
\begin{equation}
 \rho(M)=\frac{L^{-1}\sum_{\ell=1}^L E_{\ell,\mathrm{fit}}(M)}
 {L^{-1}\sum_{\ell=1}^L E_{\ell,0}(M)}.
\end{equation}
This is a ratio of layer-averaged errors. Accuracy, probability MAE, and attention MSE are macro-averaged across datasets rather than pooled across rows.

\paragraph{Implementation.}
For \method{}, key and value blocks are standardized separately before joint-KV clustering. Each layer and head uses deterministic farthest-point initialization, at most 15 Lloyd iterations, and all in-context-example query vectors for value fitting. The scalable-softmax computation retains the original example count $N$ after compression. Each head uses ridge coefficient $10^{-3}\gamma(P_{\mathrm{cal}})$ as defined in Eq.~\ref{eq:ridge-scale}.

Each layer and head stores $(\widetilde K,\widetilde V^*,\log w,N)$. Beyond the model-specific full-cache in-context-example pass, compilation costs $O(N^2d+INMd+NM^2+M^3)$ per head: full-cache target computation costs $O(N^2d)$, $I$ Lloyd iterations cost $O(INMd)$, and compressed attention plus ridge fitting costs $O(NMd+NM^2+M^3)$. Query chunking reduces peak attention workspace but not the quadratic target-computation work. Subsequent queries directly reuse the compiled memory.

% Begin sections/reproducibility
\paragraph{Dataset manifest.}
The following lists identify the 64 evaluation datasets and the ten validation datasets actually used for hyperparameter selection, not the larger pool available in the task cache.
\par\begingroup\raggedright

% Begin sections/dataset_manifest
\textbf{CC18 (64):} \path{Bioresponse}, \path{GesturePhaseSegmentationProcessed}, \path{Internet-Advertisements}, \path{MiceProtein}, \path{PhishingWebsites}, \path{adult}, \path{analcatdata_authorship}, \path{analcatdata_dmft}, \path{balance-scale}, \path{bank-marketing}, \path{banknote-authentication}, \path{blood-transfusion-service-center}, \path{breast-w}, \path{car}, \path{churn}, \path{climate-model-simulation-crashes}, \path{cmc}, \path{cnae-9}, \path{connect-4}, \path{credit-approval}, \path{credit-g}, \path{cylinder-bands}, \path{diabetes}, \path{dna}, \path{dresses-sales}, \path{electricity}, \path{eucalyptus}, \path{first-order-theorem-proving}, \path{har}, \path{ilpd}, \path{jm1}, \path{jungle_chess_2pcs_raw_endgame_complete}, \path{kc1}, \path{kc2}, \path{kr-vs-kp}, \path{madelon}, \path{mfeat-factors}, \path{mfeat-fourier}, \path{mfeat-karhunen}, \path{mfeat-morphological}, \path{mfeat-pixel}, \path{mfeat-zernike}, \path{nomao}, \path{numerai28.6}, \path{optdigits}, \path{ozone-level-8hr}, \path{pc1}, \path{pc3}, \path{pc4}, \path{pendigits}, \path{phoneme}, \path{qsar-biodeg}, \path{satimage}, \path{segment}, \path{semeion}, \path{sick}, \path{spambase}, \path{splice}, \path{steel-plates-fault}, \path{tic-tac-toe}, \path{vehicle}, \path{wall-robot-navigation}, \path{wdbc}, \path{wilt}.

\textbf{TabArena (10):} \path{APSFailure}, \path{Amazon_employee_access}, \path{Bank_Customer_Churn}, \path{Diabetes130US}, \path{MIC}, \path{SDSS17}, \path{anneal}, \path{hiva_agnostic}, \path{maternal_health_risk}, \path{students_dropout_and_academic_success}.

% End sections/dataset_manifest

\par\endgroup

\paragraph{Deterministic classification splits.}
The primary protocol uses seed 42 to create stratified 20\% test partitions and stratified in-context pools of at most 512 rows. Across the 64 datasets this yields 109,668 test rows; 13 datasets contain fewer than 512 available in-context examples. The separate repeated-split sensitivity analysis uses seeds 42--44.

\paragraph{Long-table and regression splits.}
The scale study uses adult, bank-marketing, connect-4, electricity, jungle\_chess\_2pcs\_raw\_endgame\_complete, nomao, and numerai28.6. Electricity and bank-marketing use chronological holdouts; the others use stratified holdouts. SUSY, HIGGS, and Covertype use three stratified splits. Regression uses the 13 TabArena tasks, official repeats 0--2 and fold 0, with at most 1,024 in-context examples and 200 test rows.

\paragraph{Key baseline settings.}
Input K-means, MICP, and CRUMB use standardized features and scikit-learn Lloyd K-means. CRUMB uses greedy MMD herding after grouping the unlabeled query batch. KATE and LUCoS use the released TabClustPFN \path{step-10000.ckpt} encoder; raw-space KNN uses Euclidean distance after standardization by the in-context pool.

% End sections/reproducibility

\begin{table}[H]
\centering
\caption{Paired QCOC-minus-method accuracy differences for the complete 20\% CC18 test partitions. Values are percentage points with 95\% dataset-bootstrap intervals; positive values favor QCOC.}
\label{tab:main-pairwise}
\small
\begin{tabular}{lcc}
\toprule
Method & $M=32$ & $M=64$ \\
\midrule
Random & $+17.15$ $[14.74,19.55]$ & $+11.62$ $[9.77,13.53]$ \\
Input K-means & $+12.13$ $[9.88,14.49]$ & $+6.41$ $[5.12,7.77]$ \\
LUCoS & $+11.86$ $[9.70,14.07]$ & $+6.44$ $[5.16,7.82]$ \\
MICP & $+6.29$ $[5.02,7.65]$ & $+4.29$ $[3.38,5.36]$ \\
CRUMB & $+5.01$ $[3.79,6.38]$ & $+2.70$ $[1.80,3.77]$ \\
KATE & $+4.46$ $[3.38,5.65]$ & $+2.22$ $[1.24,3.38]$ \\
Standardized raw-space KNN & $+4.17$ $[3.03,5.43]$ & $+2.11$ $[1.19,3.18]$ \\
ARASH-style & $+6.24$ $[5.11,7.44]$ & $+4.21$ $[3.23,5.32]$ \\
Weighted joint-KV & $+1.43$ $[0.86,2.20]$ & $+1.16$ $[0.58,1.94]$ \\
Full context & $-0.83$ $[-1.22,-0.48]$ & $-0.52$ $[-0.82,-0.26]$ \\
\bottomrule
\end{tabular}
\end{table}

\section{Additional Baselines and Mechanism Diagnostics}
\label{app:baselines}

\begin{table}[H]
\centering
\caption{Additional row-selection baselines on the same 64 held-out CC18 datasets under the diagnostic protocol. LUCoS uses the released TabClustPFN encoder followed by K-Medoids; raw K-Medoids uses standardized inputs. Global MMD uses one query-distribution-matched context.}
\small
\begin{tabular}{lcc}
\toprule
Method & $M=32$ & $M=64$ \\
\midrule
\multicolumn{3}{l}{\emph{Global row selection}} \\
\quad Input K-means & .7306 & .7919 \\
\quad K-Medoids (Euclidean) & .7475 & .8009 \\
\quad K-Medoids (cosine) & .7413 & .7919 \\
\quad LUCoS (Euclidean) & .7278 & .7922 \\
\quad LUCoS (cosine) & .7306 & .7934 \\
\quad MMD herding & .7509 & .8006 \\
\addlinespace[2pt]
\multicolumn{3}{l}{\emph{KV compression}} \\
\quad \textbf{\method{}} & \textbf{.8491} & \textbf{.8534} \\
\bottomrule
\end{tabular}
\end{table}

Expanding the global-selection family does not alter the main comparison: none of the representation- or input-space selectors closes the gap to \method{}. The ordering among selectors is not stable across representations, which also argues against attributing the result to one unusually weak global baseline.

\begin{table}[H]
\centering
\caption{Context reuse accounting under the diagnostic protocol. ``Passes'' is the mean number of TabICL context fits for each dataset; ``slots'' sums the in-context examples used across those fits. \method{} stores no selected raw rows and instead uses $M$ KV slots per layer and head.}
\small
\begin{tabular}{lrrrrrr}
\toprule
& \multicolumn{3}{c}{$M=32$} & \multicolumn{3}{c}{$M=64$} \\
Method & Acc. & Passes & Slots & Acc. & Passes & Slots \\
\midrule
\multicolumn{7}{l}{\emph{Dynamic row selection}} \\
\quad MICP & .7941 & 12.38 & 396 & .8181 & 6.81 & 436 \\
\quad Centroid-nearest batching & .8069 & 20 & 640 & .8313 & 20 & 1280 \\
\quad Voronoi-uniform batching & .8069 & 20 & 640 & .8369 & 20 & 1280 \\
\quad CRUMB & .8175 & 20 & 640 & .8378 & 20 & 1280 \\
\quad KATE & .8200 & 50 & 1600 & .8416 & 50 & 3200 \\
\addlinespace[2pt]
\multicolumn{7}{l}{\emph{KV compression}} \\
\quad \textbf{\method{}} & \textbf{.8491} & \textbf{1} & \textbf{0 raw} & \textbf{.8534} & \textbf{1} & \textbf{0 raw} \\
\bottomrule
\end{tabular}
\end{table}

\begin{figure}[H]
\centering
\includegraphics[width=0.94\textwidth]{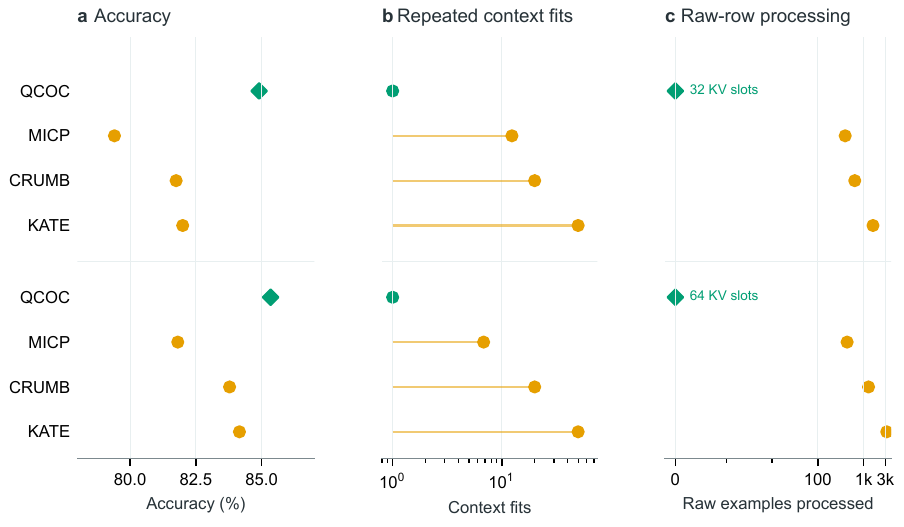}
\caption{\textbf{Accuracy and context reuse.} Rows align methods at $M=32$ and $M=64$. \textbf{A:} Mean accuracy. \textbf{B:} Mean number of context fits per dataset (log scale). \textbf{C:} Total raw in-context examples processed across those fits. QCOC processes no raw rows after compilation; its labels report the 32 or 64 retained KV slots rather than raw-row consumption.}
\label{fig:reuse-accounting}
\end{figure}

Figure~\ref{fig:reuse-accounting} shows that the retrieval baselines trade additional context fits for accuracy, whereas \method{} uses one compiled memory and remains above that frontier at both retained counts. The CRUMB variants further locate most of its improvement in query grouping rather than in replacing the selected rows alone. This supports reuse as a distinct deployment axis, rather than treating all methods with the same $M$ as operationally equivalent.

We additionally evaluate an \emph{ARASH-style} reproduction following the published locality, purity, adaptive-budget, four-way retrieval, and direct-prediction logic~\citep{arash}. The paper omits tuned thresholds and its meta-statistic rules for automatic clustering, so we fix standardized K-means, $\tau_{\mathrm{loc}}=0.5$, $\tau_{\mathrm{pur}}=0.8$, and $k_{\min}=8$. We label the result ARASH-style to distinguish it from an exact reproduction. It reaches .7953 with a mean 21.48 rows per query at maximum budget 32, and .8159 with 44.18 rows at maximum budget 64. The corresponding \method{} gains are 5.38 percentage points (95\% CI $[+4.00,+6.78]$) and 3.75 percentage points ($[+2.31,+5.16]$). We use these diagnostic results outside the primary comparison.

\begin{table}[H]
\centering
\caption{Matched compression baselines on 64 CC18 datasets at $M=32$ under the diagnostic protocol. Selection and merger choices are fixed before held-out evaluation. Intervals compare \method{} with the corresponding method.}
\label{tab:matched-compression}
\small
\begin{tabular}{lcccc}
\toprule
Method & Accuracy & Prob. MAE & Attention MSE & \method{} gain [95\% CI] \\
\midrule
\multicolumn{5}{l}{\emph{Global row selection}} \\
\quad Input K-means & .7306 & --- & --- & +.1184 [.0922,.1462] \\
\addlinespace[2pt]
\multicolumn{5}{l}{\emph{Dynamic row selection}} \\
\quad KATE & .8200 & --- & --- & +.0291 [.0137,.0444] \\
\addlinespace[2pt]
\multicolumn{5}{l}{\emph{KV compression}} \\
\quad Key centroids & .7800 & .10079 & .008627 & +.0691 [.0350,.1112] \\
\quad Key centroids + cluster size & .8259 & .05068 & .003293 & +.0231 [.0138,.0331] \\
\quad ToMe + cluster size & .8222 & .05718 & .004650 & +.0269 [.0150,.0391] \\
\quad KVMerger + cluster size & .8187 & .06258 & .005153 & +.0303 [.0172,.0438] \\
\quad Weighted joint-KV & .8316 & .04661 & .003070 & +.0175 [.0081,.0278] \\
\quad \textbf{\method{}} & \textbf{.8491} & \textbf{.01715} & \textbf{.001491} & --- \\
\midrule
Full context & .8538 & --- & --- & $-.0047$ [$-.0103$,.0003] \\
\bottomrule
\end{tabular}
\end{table}

Among matched direct-compression methods, restoring cluster mass accounts for the largest recovery over unweighted key centroids, and value fitting further reduces both fidelity errors while improving accuracy. This comparison supports the complete construction; the factorial analysis below is needed to separate the interacting roles of $N$ and cluster-count weighting.

\begin{table}[H]
\centering
\caption{\textbf{Factorial component analysis at $M=32$.} Mean results over 64 held-out CC18 datasets in the 50-query analysis setting. The last row in each block applies the same anchored value fit. ``Use original $N$'' evaluates count-dependent query transformations at the original in-context example count; crossing it out evaluates them at the compressed slot count $M$. Attention MSE measures KV reconstruction on fixed attention query vectors from the in-context examples, scaled by $N$, and therefore does not measure changes in query scaling.}
\label{tab:components}
\scriptsize
\begin{tabular}{lcccccc}
\toprule
Inputs & Use original $N$ & Cluster-count weight & Value fit & Acc. & Prob. MAE & Attention MSE \\
\midrule
Key only & $\times$ & $\times$ & $\times$ & .7581 & .1329 & .008627 \\
Key only & \checkmark & $\times$ & $\times$ & .7800 & .1008 & .008627 \\
Key only & $\times$ & \checkmark & $\times$ & .7922 & .1023 & .003293 \\
Key only & \checkmark & \checkmark & $\times$ & .8259 & .0507 & .003293 \\
Key only & \checkmark & \checkmark & \checkmark & .8469 & .0210 & .001641 \\
\addlinespace[2pt]
Joint KV & $\times$ & $\times$ & $\times$ & .7428 & .1752 & .010907 \\
Joint KV & \checkmark & $\times$ & $\times$ & .6725 & .2069 & .010907 \\
Joint KV & $\times$ & \checkmark & $\times$ & .8025 & .0967 & .003070 \\
Joint KV & \checkmark & \checkmark & $\times$ & .8316 & .0466 & .003070 \\
Joint KV & \checkmark & \checkmark & \checkmark & \textbf{.8491} & \textbf{.0171} & \textbf{.001491} \\
\bottomrule
\end{tabular}
\end{table}

The factorial design shows that preserving the original count and restoring cluster mass are complementary rather than interchangeable corrections: either one alone leaves a substantial mismatch, especially for joint-KV prototypes. After both are fixed, value fitting provides a further 1.75-point gain for joint-KV prototypes (95\% CI $[+0.81,+2.78]$). Joint-KV and key-only prototypes then have indistinguishable accuracy, but joint-KV lowers probability and attention errors on 55 and 63 of 64 datasets. The evidence therefore assigns the robust fidelity gain to joint-state construction and value fitting, while avoiding a separate accuracy claim for joint-KV clustering itself.

\begin{figure}[H]
\centering
\includegraphics[width=0.92\textwidth]{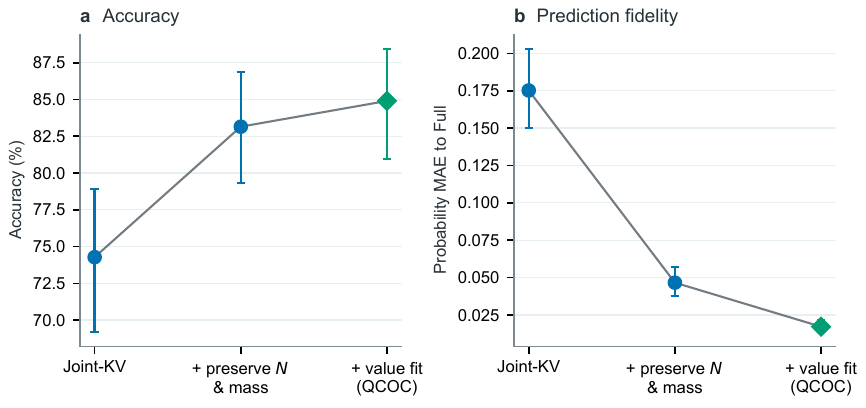}
\caption{\textbf{Component-wise accuracy and fidelity.} The stages are initial joint-KV prototypes, restoration of original $N$ and cluster-count weighting, and anchored value fitting. \textbf{A:} Mean accuracy. \textbf{B:} Mean probability MAE to Full. Error bars are dataset-bootstrap 95\% confidence intervals. Lines connect discrete constructions rather than intermediate optimization steps.}
\label{fig:component-waterfall}
\end{figure}

\section{Evaluation on Additional Models and Tasks}
\label{app:transfer}

\paragraph{TabPFN v3.}
We evaluate the pretrained TabPFN v3 version 8.2.0 on the same 64 held-out CC18 datasets with $N\leq512$, $M=32$, a requested ensemble size of one, no parameter updates, and no additional tuning. TabPFN's default feature-coverage rule automatically increases the ensemble size on high-dimensional datasets; this behavior is shared by all methods. Row-selection baselines retain the same examples selected in the TabICL experiment and delegate prediction to TabPFN, with KATE fitting a separate context for each query. When a KATE context contains no usable nonconstant feature and TabPFN rejects it, we return that context's empirical class probabilities. For the direct baselines, we dequantize the explicit int8 cache and apply each compression method to the same full cache. For \method{}, compressed values are shared across the eight query heads associated with each KV head, and the compressed cache carries the original in-context example count into any active softmax-scaling layer.

\begin{table}[H]
\centering
\caption{\textbf{Matched baselines on TabPFN v3.} Probability MAE and attention-output MSE apply only to direct cache-compression methods and measure error relative to dequantized full-context inference. The final column gives the paired QCOC-minus-baseline accuracy difference with a 95\% dataset-bootstrap interval.}
\label{tab:tabpfn-transfer}
\scriptsize
\begin{tabular}{lcccc}
\toprule
Method & Accuracy & Prob. MAE $\downarrow$ & Attention-output MSE $\downarrow$ & $\Delta$ accuracy \\
\midrule
Class-balanced random & .6922 & --- & --- & +.1528 [.1234,.1841] \\
Input K-means & .7184 & --- & --- & +.1266 [.0937,.1613] \\
LUCoS (Euclidean) & .7244 & --- & --- & +.1206 [.0897,.1525] \\
LUCoS (cosine) & .7366 & --- & --- & +.1084 [.0784,.1406] \\
MICP & .7966 & --- & --- & +.0484 [.0312,.0659] \\
CRUMB & .8263 & --- & --- & +.0187 [.0066,.0312] \\
KATE & .8200 & --- & --- & +.0250 [.0094,.0425] \\
\midrule
Key centroids & .8350 & .03883 & .014916 & +.0100 [.0019,.0184] \\
Weighted key K-means & .8353 & .03397 & .007718 & +.0097 [.0022,.0181] \\
Weighted ToMe & .8319 & .04053 & .009107 & +.0131 [.0053,.0219] \\
Weighted KVMerger & .8341 & .04474 & .009881 & +.0109 [.0028,.0197] \\
Weighted joint-KV & .8397 & .03667 & .007234 & +.0053 [$-.0019$,.0125] \\
\textbf{\method{}} & \textbf{.8450} & \textbf{.01414} & \textbf{.004131} & --- \\
Full context & .8509 & --- & --- & $-.0059$ [$-.0109$,$-.0012$] \\
\bottomrule
\end{tabular}
\end{table}

The transfer experiment reproduces the same qualitative ordering on a different prediction model: \method{} separates clearly from row selection and from uncalibrated cache compression, while remaining close to Full. Its gain over weighted joint-KV is small and statistically unresolved, but its probability and attention errors are substantially lower. Thus the transferable result is the benefit of calibrating compressed values for full-cache fidelity; the experiment does not establish that joint-KV clustering alone improves TabPFN accuracy.

\begin{figure}[H]
\centering
\includegraphics[width=0.96\textwidth]{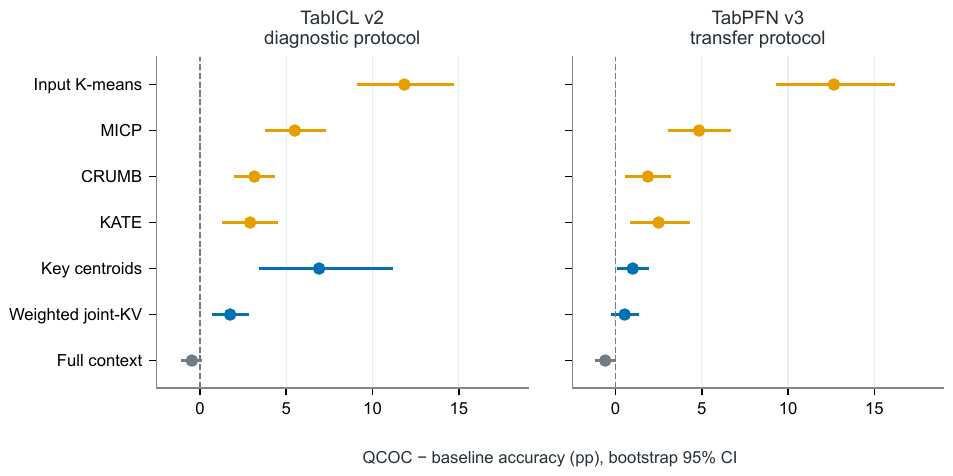}
\caption{\textbf{Cross-model accuracy comparison.} Points show mean QCOC-minus-baseline accuracy differences and bars show 95\% dataset-bootstrap intervals over the same 64 CC18 datasets. TabICL uses the diagnostic protocol and TabPFN the transfer protocol; both panels share the same horizontal scale.}
\label{fig:cross-model-forest}
\end{figure}

\subsection{Regression}
\label{app:regression}

We evaluate TabICL regression on all 13 TabArena-v0.1 regression datasets, using official OpenML repeats 0--2 and fold 0. Each split uses at most 1,024 in-context examples and 200 test rows. Lower normalized RMSE is better.

\begin{table}[H]
\centering
\caption{TabArena regression over 13 datasets and three official splits per dataset. Lower normalized RMSE and higher $R^2$ are better; bold marks the best compressed or retrieval method at each retained count. KATE uses the same encoder-space retrieval protocol as in classification.}
\small
\begin{tabular}{clcc}
\toprule
Budget & Method & Norm. RMSE $\downarrow$ & $R^2\uparrow$ \\
\midrule
\multirow{4}{*}{32} & Full context & .4582 & .7398 \\
 & KATE & .5261 & .6728 \\
 & Weighted joint-KV & .5796 & .6204 \\
 & \method{} & \textbf{.5127} & \textbf{.6864} \\
\midrule
\multirow{4}{*}{64} & Full context & .4582 & .7398 \\
 & KATE & .4946 & .7074 \\
 & Weighted joint-KV & .5488 & .6515 \\
 & \method{} & \textbf{.4883} & \textbf{.7101} \\
\bottomrule
\end{tabular}
\end{table}

On regression, \method{} consistently recovers much of the degradation introduced by direct weighted-KV compression at both retained counts. Its mean normalized RMSE is also lower than KATE's, but those differences are statistically unresolved. The regression evidence therefore supports transfer beyond classification for direct cache compression, without claiming superiority over retrieval on this smaller benchmark.

\section{Iterative End-to-End Refinement}
\label{app:e2e}

To test whether the layerwise closed-form attention-output match can be improved through the final prediction objective, we evaluate an iterative end-to-end extension. Starting from the $M=32$ \method{} memory, we keep the model parameters, compressed keys, and cluster-count weights unchanged and update only the compressed values. Using 128 independent unlabeled calibration rows, we take centered logits from the full cache as teacher targets and minimize centered-logit MSE together with an anchor penalty to the closed-form solution. The learning rate, number of steps, and anchor weight are selected on three validation datasets and then fixed to $3\times10^{-4}$, 30, and $10^{-4}$, respectively.

Across ten validation datasets, iterative refinement increases mean accuracy from 88.60\% to 89.20\%, with a 2/8/0 win/tie/loss record; probability MAE decreases from 0.0201 to 0.0153, and centered-logit MSE from 0.3227 to 0.0442. The fidelity improvement does not consistently translate into higher long-context accuracy. On \texttt{electricity}, at $N=8{,}192$, probability MAE decreases from 0.1543 to 0.1076 while accuracy drops from 0.80 to 0.78; at $N=32{,}768$, probability MAE decreases from 0.2485 to 0.2029 and accuracy rises from 0.80 to 0.82, still leaving a gap to full context. The extension also introduces extra unlabeled data, backpropagation, and iterative hyperparameters. We therefore retain it as a fidelity diagnostic and optional extension, while the main paper uses the closed-form method that requires neither an additional calibration set nor gradient-based optimization.

\section{Long-Table Results}
\label{app:long}

\begin{table}[H]
\centering
\caption{Results for the seven CC18 long-table datasets listed in Appendix~\ref{app:details}. \texttt{electricity} and \texttt{bank-marketing} use chronological holdouts; the other five use fixed stratified splits. All methods use 8,192 in-context examples and 1,000 test rows; compressed methods use $M=512$.}
\label{tab:long-results}
\small
\resizebox{\textwidth}{!}{%
\begin{tabular}{lccccccc}
\toprule
Dataset & Full & Input K-means & MICP & CRUMB & KATE & Weighted KV & \method{} \\
\midrule
adult & .871 & .841 & .850 & .863 & .859 & .865 & .870 \\
bank-marketing & .758 & .740 & .748 & .739 & .737 & .763 & .754 \\
connect-4 & .810 & .734 & .779 & .770 & .793 & .762 & .807 \\
electricity & .878 & .880 & .854 & .869 & .883 & .880 & .871 \\
jungle chess & .878 & .809 & .868 & .885 & .884 & .868 & .879 \\
nomao & .968 & .943 & .954 & .960 & .971 & .959 & .971 \\
numerai28.6 & .491 & .508 & .498 & .497 & .486 & .488 & .505 \\
\midrule
Macro & .8077 & .7793 & .7930 & .7976 & .8019 & .7979 & .8081 \\
\bottomrule
\end{tabular}}
\end{table}

\begin{table}[H]
\centering
\caption{\textbf{Long-context accuracy and cache reduction.} All results use $N=8{,}192$ and $M=512$; external-dataset accuracies average three splits. Cache reduction compares aggregate cache bytes with Full, not memory use against row-selection methods. Bold marks the best compressed method.}
\label{tab:long-external}
\small
\resizebox{\textwidth}{!}{%
\begin{tabular}{lccccccc}
\toprule
Benchmark & Full & MICP & CRUMB & KATE & Weighted KV & \method{} & Cache reduction \\
\midrule
CC18-Long (7 datasets) & .8077 & .7930 & .7976 & .8019 & .7979 & \textbf{.8081} & 10.53$\times$ \\
SUSY & .8067 & .7930 & .7890 & .7893 & .7953 & \textbf{.8077} & 12.09$\times$ \\
HIGGS & .7167 & .6803 & .6853 & .7007 & .7090 & \textbf{.7110} & 10.94$\times$ \\
Covertype & .8580 & .8130 & .8157 & \textbf{.8530} & .7707 & .8153 & 9.03$\times$ \\
\bottomrule
\end{tabular}}
\end{table}

\begin{table}[H]
\centering
\caption{Memory-budget sweep on the same seven long tables. Cache compression is the ratio of aggregate full-cache bytes to aggregate \method{} cache bytes; it is not a comparison with row-selection memory.}
\small
\begin{tabular}{rcccc}
\toprule
$M$ & Accuracy & Gap to Full & Agreement & Cache compression \\
\midrule
256 & .8056 & $-.0021$ & .9626 & 15.50$\times$ \\
512 & .8081 & $+.0004$ & .9751 & 10.53$\times$ \\
1,024 & .8069 & $-.0009$ & .9780 & 6.42$\times$ \\
\bottomrule
\end{tabular}
\end{table}

Across the seven long tables, value fitting sharply reduces probability error and raises mean accuracy, although the accuracy interval remains inconclusive at this sample size. The budget sweep is non-monotonic: increasing $M$ beyond 512 improves agreement with Full but not task accuracy. More slots therefore improve fidelity without necessarily correcting the remaining classification errors, and $M=512$ gives the best observed accuracy--memory compromise.

\section{In-Context Example/Memory Settings, Serving, and Compilation Time}
\label{app:scale-matrix}

Table~\ref{tab:scale-latency} complements the accuracy matrix in
Table~\ref{tab:scale-matrix} with single-core CPU serving time. Selection,
embedding, clustering, and \method{} compilation are excluded. Fixed row
baselines fit one selected context and process all 1,000 queries as a batch;
MICP and CRUMB fit their routed contexts and batch the queries assigned to each;
KATE and ARASH-style execute query-specific contexts; and KV methods reuse a
prebuilt cache for one 1,000-query batch. Context-method timings are medians of
three independent runs, while Full and KV methods average ten warmed predictions.
To control CPU parallelism, we measure compilation and inference wall-clock time using one thread on an Intel Xeon 6767P CPU. Timing runs match the in-context example count, retained count, and query count of the accuracy evaluation.

\begin{table}[t]
\centering
\caption{\textbf{Single-core CPU serving milliseconds per query across all in-context example count and retained-count settings.} Means are over the same seven datasets and input-size settings as Table~\ref{tab:scale-matrix}. Rows are methods with a complete matched CPU serving sweep; each $N$ block uses $M=N/32,N/16,N/8$ from left to right. Values use each method's intended per-query, routed-batch, or fixed-batch execution and exclude selection and compilation.}
\label{tab:scale-latency}
\scriptsize
\setlength{\tabcolsep}{2.7pt}
\resizebox{\textwidth}{!}{%
\begin{tabular}{lccc@{\hspace{5pt}}ccc@{\hspace{5pt}}ccc@{\hspace{5pt}}ccc}
\toprule
& \multicolumn{3}{c}{$N=512$} & \multicolumn{3}{c}{$N=2{,}048$} & \multicolumn{3}{c}{$N=4{,}096$} & \multicolumn{3}{c}{$N=8{,}192$} \\
\cmidrule(lr){2-4}\cmidrule(lr){5-7}\cmidrule(lr){8-10}\cmidrule(lr){11-13}
Method & 16 & 32 & 64 & 64 & 128 & 256 & 128 & 256 & 512 & 256 & 512 & 1,024 \\
\midrule
\multicolumn{13}{l}{\emph{Global row selection}} \\
\quad Random & 1.83 & 2.05 & 2.14 & 2.18 & 2.36 & 2.69 & 2.38 & 2.76 & 3.57 & 2.76 & 3.51 & 5.21 \\
\quad Input K-means & 2.06 & 2.15 & 2.21 & 2.27 & 2.42 & 2.75 & 2.44 & 2.73 & 3.50 & 2.82 & 3.59 & 4.92 \\
\quad LUCoS & 1.96 & 2.05 & 2.18 & 2.15 & 2.42 & 2.78 & 2.42 & 2.68 & 3.42 & 2.77 & 3.59 & 4.94 \\
\addlinespace[2pt]
\multicolumn{13}{l}{\emph{Dynamic row selection}} \\
\quad MICP & 3.50 & 3.01 & 2.71 & 6.26 & 5.99 & 5.91 & 9.69 & 9.52 & 10.30 & 16.88 & 18.47 & 18.84 \\
\quad CRUMB & 3.31 & 4.40 & 5.76 & 5.14 & 7.76 & 13.53 & 7.69 & 12.94 & 26.36 & 12.32 & 25.88 & 55.68 \\
\quad KATE & 77.98 & 113.54 & 179.58 & 153.57 & 263.09 & 503.66 & 252.17 & 478.20 & 997.97 & 453.82 & 971.21 & 2296.13 \\
\quad ARASH-style & 76.29 & 90.35 & 123.18 & 121.45 & 199.26 & 324.15 & 171.34 & 318.77 & 598.07 & 284.68 & 531.65 & 1332.73 \\
\addlinespace[2pt]
\multicolumn{13}{l}{\emph{KV compression}} \\
\quad Weighted joint-KV & 1.94 & 1.95 & 1.97 & 1.96 & 1.98 & 2.01 & 1.87 & 1.93 & 2.00 & 1.92 & 1.94 & 2.15 \\
\quad \method{} & 1.93 & 1.99 & 1.96 & 1.80 & 1.90 & 2.00 & 1.81 & 1.70 & 2.12 & 1.72 & 1.91 & 2.33 \\
\addlinespace[2pt]
Full context & 2.06 & 2.06 & 2.06 & 2.48 & 2.53 & 2.47 & 2.89 & 2.89 & 2.87 & 3.85 & 3.79 & 3.86 \\
\bottomrule
\end{tabular}}
\end{table}

The serving curves have qualitatively different scaling. Because \method{} reuses a fixed $M$-slot cache, its online cost is nearly flat as the source context grows; Full continues to attend over $N$, while query-specific retrieval repeatedly constructs and executes a new context. The online advantage therefore comes from reuse and bounded attention length together, not from compression ratio alone. These measurements exclude one-time setup cost.

\begin{table}[htbp]
\centering
\caption{\textbf{QCOC single-core CPU compilation time (seconds).} One run on the \texttt{adult} dataset per setting on an Intel Xeon 6767P, with $M=N/16$. Times begin from the constructed full cache and include prototype construction, count estimation, and value fitting.}
\label{tab:compile-time}
\footnotesize
\begin{tabular}{lrrrr}
\toprule
Method & $512/32$ & $2{,}048/128$ & $4{,}096/256$ & $8{,}192/512$ \\
\midrule
\method{} & 1.86 & 21.01 & 74.36 & 283.40 \\
\bottomrule
\end{tabular}
\end{table}

Compilation grows steeply with context length and dominates the cost at the largest setting. Tables~\ref{tab:scale-latency} and~\ref{tab:compile-time} therefore describe opposite sides of the deployment tradeoff: a substantial one-time investment buys a much smaller reusable online state. Whether that investment is worthwhile depends on query volume.

\FloatBarrier
\setlength{\textfloatsep}{8pt plus 2pt minus 2pt}
\setlength{\intextsep}{8pt plus 2pt minus 2pt}
\section{External Long-Table Details}
\label{app:external-long}

The residual analysis asks whether a label-free reconstruction signal can expose failure of one shared memory. Across the ten long tables, larger attention-output residuals tend to coincide with larger losses relative to Full ($r=0.74$). This association is useful as a diagnostic hypothesis, but the sample is too small to define a calibrated fallback threshold.

\begin{table}[htbp]
\centering
\caption{\textbf{External long-table accuracy by split.} All runs use $N=8{,}192$, $M=512$, and 1,000 test queries.}
\label{tab:external-splits}
\footnotesize
\setlength{\tabcolsep}{3.5pt}
\begin{tabular}{llcccccc}
\toprule
Dataset & Split & Full & \method{} & KATE & Weighted KV & CRUMB & MICP \\
\midrule
\multirow{4}{*}{SUSY}
 & 1 & .797 & .797 & .773 & .774 & .779 & .780 \\
 & 2 & .813 & .813 & .797 & .812 & .788 & .795 \\
 & 3 & .810 & .813 & .798 & .800 & .800 & .804 \\
 & Mean & .8067 & \textbf{.8077} & .7893 & .7953 & .7890 & .7930 \\
\midrule
\multirow{4}{*}{Covertype}
 & 1 & .855 & .821 & .854 & .782 & .823 & .815 \\
 & 2 & .885 & .835 & .873 & .788 & .835 & .827 \\
 & 3 & .834 & .790 & .832 & .742 & .789 & .797 \\
 & Mean & .8580 & .8153 & \textbf{.8530} & .7707 & .8157 & .8130 \\
\midrule
\multirow{4}{*}{HIGGS}
 & 1 & .727 & .724 & .724 & .727 & .695 & .698 \\
 & 2 & .698 & .693 & .668 & .680 & .663 & .652 \\
 & 3 & .725 & .716 & .710 & .720 & .698 & .691 \\
 & Mean & .7167 & \textbf{.7110} & .7007 & .7090 & .6853 & .6803 \\
\bottomrule
\end{tabular}
\end{table}

The split-level results show two distinct behaviors rather than one noisy average. QCOC remains close to Full and matches or exceeds the retrieval baselines on all SUSY and HIGGS splits, whereas KATE consistently approaches Full more closely on Covertype. The Covertype gap is therefore a repeatable distribution-level failure of one shared memory, not an artifact of a single split.

\begin{table}[htbp]
\centering
\caption{\textbf{Covertype budget sensitivity.} Means over three splits at $N=8{,}192$. Agreement is measured against Full, and $\rho(M)$ is the normalized attention-output residual defined in Appendix~\ref{app:details}.}
\label{tab:covertype-adaptive}
\footnotesize
\begin{tabular}{rcccc}
\toprule
$M$ & Accuracy & Gap to Full & Agreement & $\rho(M)$ \\
\midrule
512 & .8153 & $-4.27$ pt & .8973 & .511 \\
1,024 & .8317 & $-2.63$ pt & .9357 & .467 \\
\bottomrule
\end{tabular}
\end{table}

Doubling the retained count reduces both the residual and the accuracy gap on Covertype, so the diagnostic moves in the expected direction. It recovers only part of the loss, however, showing that additional capacity mitigates but does not remove the mismatch between a single shared memory and this dataset's local structure.

\begin{figure}[!htbp]
\centering
\includegraphics[width=0.55\textwidth]{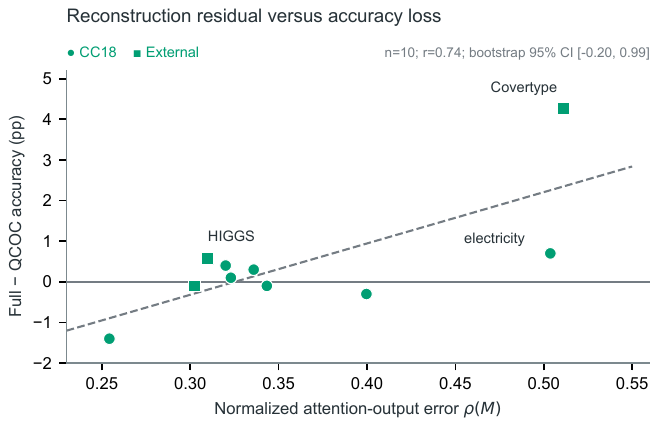}
\caption{\textbf{Reconstruction residual and accuracy loss.} Each point is one QCOC-compressed long table; circles denote seven CC18 tables and squares three external tables. The dashed line is a descriptive least-squares fit ($r=0.74$; dataset-bootstrap 95\% confidence interval $[-0.20,0.99]$), not a confidence band or a calibrated decision rule.}
\label{fig:residual-diagnostic}
\end{figure}

% End sections/appendix

\end{document}